\documentclass[10pt,twocolumn,letterpaper]{article}
\usepackage{iccv}              

\usepackage{subcaption}
\usepackage{amsmath,amssymb,amsfonts}
\usepackage[accsupp]{axessibility}
\usepackage{multirow}
\usepackage{graphicx}
\usepackage{makecell}
\usepackage{textcomp}
\usepackage{algorithm}
\usepackage{algpseudocode}
\usepackage{xcolor}
\usepackage[labelformat=simple, labelsep=period]{subcaption}

\def\BibTeX{{\rm B\kern-.05em{\sc i\kern-.025em b}\kern-.08em
    T\kern-.1667em\lower.7ex\hbox{E}\kern-.125emX}}

\definecolor{iccvblue}{rgb}{0.21,0.49,0.74}
\usepackage[pagebackref,breaklinks,colorlinks,allcolors=iccvblue]{hyperref}

\def\paperID{3} 
\def\confName{ICCV}
\def\confYear{2025}

\title{Fairness Without Labels: Pseudo-Balancing for Bias Mitigation in Face Gender Classification}

\author{
Haohua Dong$^{1,2}$ \quad Ana Manzano Rodríguez$^{1,3}$ \quad Camille Guinaudeau$^{4}$ \quad Shin’ichi Satoh$^{1}$ \\
$^1$National Institute of Informatics, Japan \\
$^2$INESC-ID, Instituto Superior Técnico, University of Lisbon, Portugal \\
$^3$University of Amsterdam, The Netherlands \\
$^4$LIMSI, CNRS / Université Paris-Saclay, France \\
{\tt\small haohua@tecnico.ulisboa.pt}
}

\begin{document}
\maketitle
\begin{abstract}
Face gender classification models often reflect and amplify demographic biases present in their training data, leading to uneven performance across gender and racial subgroups. We introduce \textit{pseudo-balancing}, a simple and effective strategy for mitigating such biases in semi-supervised learning. Our method enforces demographic balance during pseudo-label selection, using only unlabeled images from a race-balanced dataset without requiring access to ground-truth annotations. 

We evaluate pseudo-balancing under two conditions: (1) fine-tuning a biased gender classifier using unlabeled images from the FairFace dataset, and (2) stress-testing the method with intentionally imbalanced training data to simulate controlled bias scenarios. In both cases, models are evaluated on the All-Age-Faces (AAF) benchmark, which contains a predominantly East Asian population. Our results show that pseudo-balancing consistently improves fairness while preserving or enhancing accuracy. The method achieves 79.81\% overall accuracy—a 6.53\% improvement over the baseline—and reduces the gender accuracy gap by 44.17\%. In the East Asian subgroup, where baseline disparities exceeded 49\%, the gap is narrowed to just 5.01\%. 
These findings suggest that even in the absence of label supervision, access to a demographically balanced or moderately skewed unlabeled dataset can serve as a powerful resource for debiasing existing computer vision models.
\end{abstract}    
\section{Introduction}

Automated gender classification systems are widely deployed in modern computer vision applications, yet their performance disparities across demographic groups remain a persistent challenge. While these systems achieve high accuracy on Western populations, their reliability degrades significantly when applied to non-Western population such as Black or East Asian faces. This performance disparity often stems from two compounding factors: biased training datasets that under-represent non-Western populations \cite{huang_face_data_set_2008_10, Phillips1998TheFD_11}, and algorithmic approaches that amplify these biases during learning \cite{KRISHNAN2023_new_bias, gender_shades_2018}.

Balanced training datasets like FairFace (FF) \cite{karkkainenfairface} demonstrate that careful curation can improve fairness, but recollecting such data is often prohibitively expensive. Semi-supervised learning offers a cost-effective alternative, but standard pseudo-labeling approaches like FixMatch \cite{sohn2020fixmatchsimplifyingsemisupervisedlearning} often reinforce existing biases rather than mitigate them, as the model is trained on pseudo-labels it generates itself—thus amplifying its initial bias.

In this work, we explore how unlabeled images from a race-balanced dataset like FairFace can be strategically used to reduce bias—without relying on FairFace's ground truth labels. We introduce pseudo-balancing, a lightweight yet effective enhancement to self-training pipelines, which dynamically adjusts pseudo-label selection to enforce gender balance.
Our method (1) modifies only the sample selection step and prevents majority-class domination during self-training by manually re-balancing pseudo-labels, (2) preserves model accuracy through confidence-based sample selection, and (3) avoids the need for adversarial training or distributional assumptions common in domain adaptation techniques \cite{ganin2016domainadversarialtrainingneuralnetworks}. This makes our approach highly compatible with existing semi-supervised methods such as FixMatch \cite{sohn2020fixmatchsimplifyingsemisupervisedlearning} and FlexMatch \cite{zhang2022flexmatchboostingsemisupervisedlearning}, which leverage confident pseudo-labeling and adaptive thresholding strategies to improve learning from unlabeled data. 
We structure our study around two experimental scenarios designed to evaluate both the benefits and the limits of this approach:
\begin{itemize}
    \item \textbf{Scenario 1}: We start with a standard case where a classifier is trained on a publicly available but demographically skewed dataset Kaggle Gender Classification Dataset \cite{kaggle}. We then apply pseudo-balancing using unlabeled FF images to rebalance the training distribution. Evaluation is conducted on the All-Age-Faces (AAF) dataset \cite{cheng2019exploiting_aaf}, which contains predominantly East Asian images. This scenario tests whether incorporating balanced unlabeled data during training can improve fairness over a biased baseline;
    
    \item \textbf{Scenario 2}: We simulate controlled bias conditions by training the initial classifier on data from only varying demographic group imbalances (e.g., gender and race subsets from FF). We then apply pseudo-balancing using unlabeled FF images. This more challenging setup evaluates the robustness of our method when the labeled training data lacks diversity entirely.
\end{itemize}
In both scenarios, we evaluate on the AAF benchmark using overall accuracy and selection rate (accuracy gap across genders). Notably, even without FairFace labels, leveraging its balanced composition reduces gender disparities. In Scenario 1, our method achieves a more uniform accuracy than a Kaggle-trained baseline. In Scenario 2, pseudo-balancing remains effective despite initial bias in training data, demonstrating its robustness.

\noindent \textbf{Our main contributions are as follows:}
\begin{itemize}
\item We propose pseudo-balancing, a simple bias mitigation method that enforces pseudo-gender balance during self-training without requiring labeled source data;
\item We introduce a framework to evaluate compounding dataset and model biases via controlled experiments using FairFace and All-Age-Faces;
\item We empirically show that pseudo-balancing reduces gender disparities while maintaining or improving accuracy on the AAF benchmark.
\end{itemize}

Our method achieves 79.81\% accuracy—a 6.53\% gain over the baseline—while cutting the gender gap by 44.17\%. Pseudo-balancing particularly benefits underrepresented groups in the training data, offering a fairer gender classification without costly labeled data collection. 

The rest of the paper is organized as follows: Section~\ref{sec:related_work} reviews related work. Section~\ref{sec:methods} details our approach. Section~\ref{sec:experiments} presents experiments. Section~\ref{sec:discussion} discusses implications, with limitations and conclusions in Sections~\ref{sec:limitations} and~\ref{sec:conclusion}.

\section{Related Work}
\label{sec:related_work}

\subsection{Gender Classification Algorithms: Its Applications and Bias}
Automated gender classification systems is widely used across multiple domains, including video surveillance \cite{Demirkus2010AutomatedPC_3}, demographic analysis \cite{hu_2014_estimation_face_12}, targeted advertising \cite{sandhu_gender_advertisement_2018_4}, and human-computer interaction \cite{comparative_gender_classification_2014_5}. However, extensive research has revealed significant performance disparities in commercial systems, particularly for non-Western populations such as East Asians or Black people. For example, in a 2018 study, a dermatologist approved Fitzpatrick skin type classification system was shown to be have error rates as high as 34.7\% for dark-skinned females compared to light-skinned males \cite{gender_shades_2018}.

These technical shortcomings have real-world consequences, especially in high-stakes applications. Law enforcement systems using biased facial recognition have been shown to subject African-American individuals to searches at disproportionate rates \cite{Garvie2016_21}, while commercial systems frequently reinforce harmful stereotypes by associating women with domestic activities \cite{zhao2017menlikeshoppingreducing_9} and being misgendered by automatic gender recognition systems (AGR) could lead to a greater insult due to the perceived objectivity of computer systems \cite{kumar_2016_face_recognition_trans_17}. The root causes of these biases can often be traced to unrepresentative training datasets.

For example, Color FERET is a 8.5GB database benchmark for facial recognition under controlled environments \cite{Phillips1998TheFD_11}. Color FERET contains only 10\% as many African male identities and 17\% as many African female identities compared to their white counterparts. Labeled Faces in The Wild (LFW), a database composed of celebrity facial images designed for facial verification \cite{huang_face_data_set_2008_10}, is estimated to be composed of 77.5\% male and 83.5\% white faces \cite{hu_2014_estimation_face_12}. Recent facial recognition systems reported a 97.35\% accuracy on the LFW database, however it is not clear how the algorithm is performing on the underrepresented subgroups inside the benchmark \cite{gender_shades_2018, taigman_2014_13}.

Recent initiatives have attempted to address these representation gaps through improved dataset design. The Diversity in Faces (DiF) benchmark \cite{merler2019diversityfaces_15} provides comprehensive anthropometric measurements across one million images, while FairFace (FF) \cite{karkkainenfairface} explicitly balances representation across seven racial groups (White, Black, Indian, East Asian, Southeast Asian, Middle Eastern, and Latino) with near-equal gender distribution. These datasets demonstrate that balanced training data can improve model fairness. However, even carefully curated datasets struggle with intersectional biases (e.g., dark-skinned women) \cite{KRISHNAN2023_new_bias} and require complementary algorithmic solutions.

\begin{figure}[h]
    \centering
    \begin{subfigure}{0.4\linewidth}
        \centering
        \includegraphics[width=\linewidth]{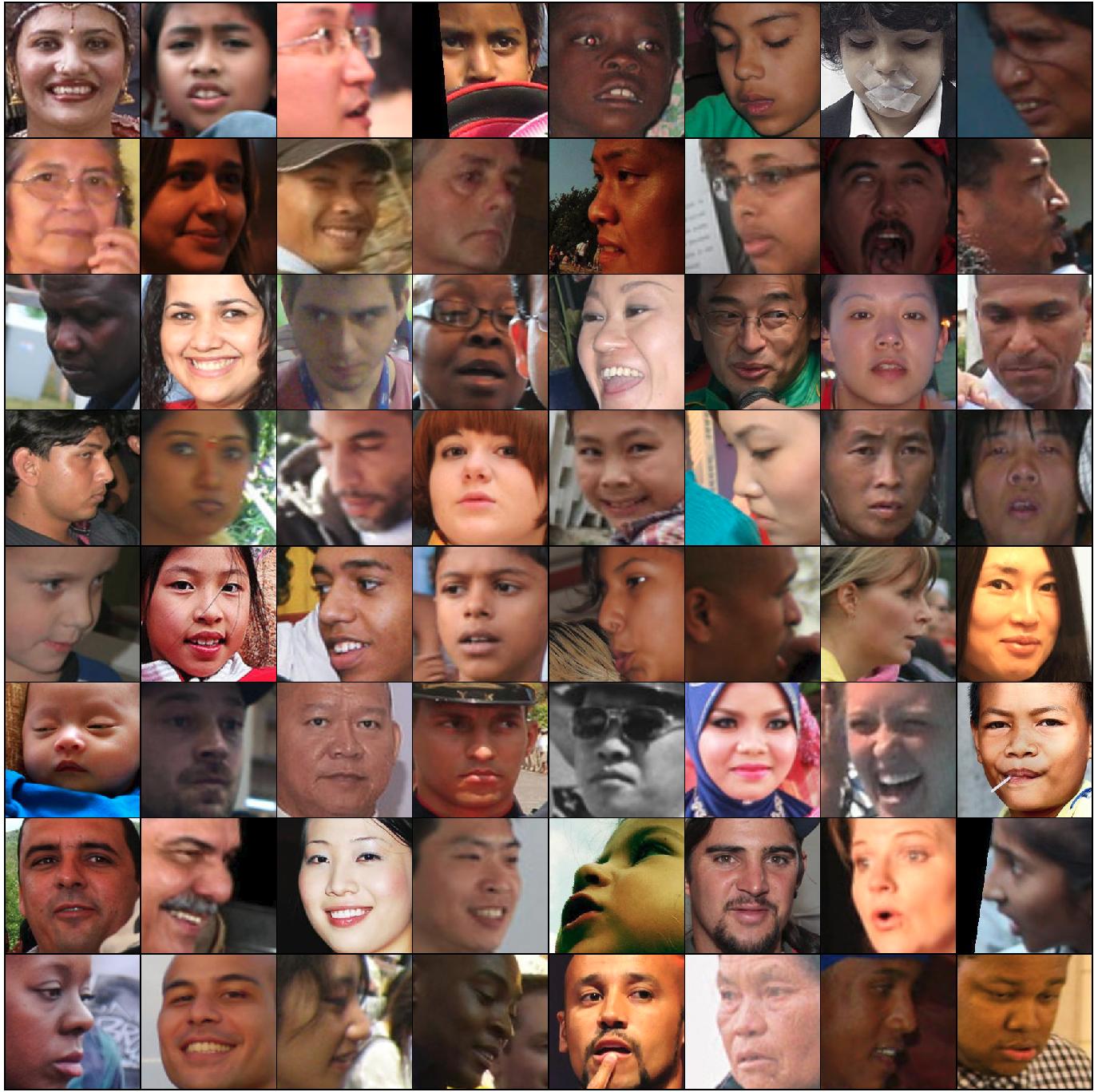}
        \caption{FairFace (FF) dataset}
        \label{fig:fairface}
    \end{subfigure}
    \hfill
    \begin{subfigure}{0.49\linewidth}
        \centering
        \includegraphics[width=\linewidth]{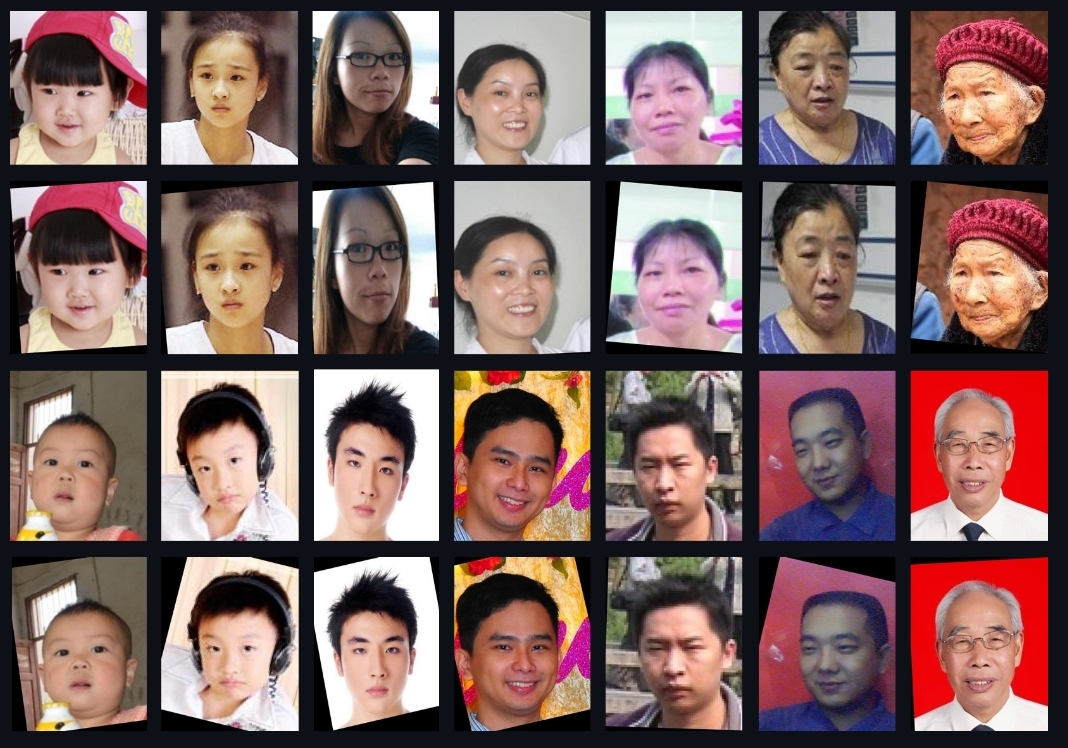}
        \caption{All-Ages-Face (AAF) dataset}
        \label{fig:aaf}
    \end{subfigure}
    \caption{FairFace (FF) and All-Ages-Face (AAF) dataset examples.}
    \label{fig:dataset-comparison}
    \end{figure}
    
\subsection{Traditional Approaches for Fairness Improvement}
While our previous discussion highlighted how skewed training data distributions propagate bias, the machine learning architectures themselves can introduce and amplify these disparities. Traditional approaches to improving fairness involve fine-tuning models with ground truth labels, which, while effective, require extensive manual annotation \cite{hoffman2017cycadacycleconsistentadversarialdomain}. Another strategy involves retraining models on more balanced datasets, though this is resource-intensive and may still be affected by the biases in the new dataset \cite{ganin2016domainadversarialtrainingneuralnetworks, KRISHNAN2023_new_bias}. This motivates our investigation of alternative solutions, particularly for applications like surveillance and targeted advertising where annotation budgets are limited.

To address these challenges, semi-supervised and unsupervised learning techniques have been proposed. Self-supervised pre-training methods like MoCo \cite{he2019moco, chen2020mocov2} and SimCLR \cite{chen2020simpleframeworkcontrastivelearning} offer another pathway by learning robust representations from uncurated data. However, such approaches still risk encoding societal biases present in the pre-training corpus \cite{zhao2017menlikeshoppingreducing_9}. Domain-Adversarial Neural Networks (DANN) \cite{ganin2016domainadversarialtrainingneuralnetworks} leverage adversarial learning to mitigate domain shifts and reduce bias, making them promising candidates for fairer gender classification, but require careful tuning to handle demographic imbalances.
Pseudo-labeling techniques have emerged as effective approaches for finetuning pre-trained models, generating synthetic labels from unlabeled data to improve feature alignment across demographic groups. Recent advances in curriculum-based methods like FlexMatch \cite{zhang2022flexmatchboostingsemisupervisedlearning} demonstrate how adaptive thresholding - dynamically adjusting confidence requirements based on model performance - can outperform fixed-threshold approaches like FixMatch \cite{sohn2020fixmatchsimplifyingsemisupervisedlearning} in biased learning scenarios. Building on these foundations, contemporary research highlights the importance of balanced pseudo-label distributions \cite{he2021redistributingbiasedpseudolabels}, suggesting that careful management of demographic representation during training can further enhance model fairness.

This critical gap in existing approaches motivates our comprehensive study of bias mitigation techniques using FairFace's carefully balanced demographic structure as both an experimental framework and source domain. FairFace's controlled composition (equal representation across seven racial groups and near-balanced gender distribution) provides an interesting testbed for evaluating algorithmic fairness interventions. Our experimental design isolates and analyzes specific bias factors - particularly the intersection of racial and gender biases that affect East Asian populations in the All-Ages-Faces dataset \cite{cheng2019exploiting_aaf}.

\section{Methods}
\label{sec:methods}
\subsection{Model and Datasets}

We employ a readily available ResNet18-based convolutional neural network \cite{he2015deepresiduallearningimage} pre-trained on the Kaggle gender classification dataset as our baseline model\footnote{https://github.com/ndb796/Face-Gender-Classification-PyTorch}. This choice is motivated by its demonstrated effectiveness when fine-tuned on FairFace and mixed datasets, achieving a significant reduction in the gender accuracy gap for Japanese data \cite{manzano2025genderbias}, despite its known limitations in performance on minority subgroups (See Table \ref{tab:previous_results}). The model is fine-tuned using FairFace \cite{karkkainenfairface}, All-Ages-Face (AAF) datasets \cite{cheng2019exploiting_aaf}, and Kaggle Gender Classification Dataset \cite{kaggle}. FairFace contains 108k training images balanced across seven racial groups (with a rate of 53\% male, 47\% female), AAF comprises 13.3k images primarily featuring individuals of Asian descent (45\% male, 55\% female), while Kaggle contains 47k training images (49.5\% female, 50.5\% male). 

We measure bias using the \textbf{Selection Rate (SR)}, defined as the ratio of the minimum to maximum gender accuracy:
\[
\text{SR} = \frac{\text{min accuracy}}{\text{max accuracy}}.
\]
Following prior work~\cite{evaluatiob_metric_2015, gender_classification_bias_2020_1}, SR~$\geq$~80\% indicates mitigated bias. Bias reduction is considered effective when SR increases along with overall model accuracy.

\subsection{Supervised Learning}
We conducted fine-tuning experiments in a supervised learning setting \cite{manzano2025genderbias}, using a gender-balanced test set of 2,504 samples from the All-Ages-Face (AAF) dataset to ensure fair evaluation across male and female subgroups. Fine-tuning the baseline model on the FairFace dataset improved overall accuracy on AAF but left a significant gender accuracy gap of 26\%—a reduction from the original 49\%. The most equitable performance was achieved by sequentially fine-tuning on FairFace and then a mixed dataset (AAF and Kaggle Gender Classification), reducing the gender disparity to under 1\%, as shown in Table~\ref{tab:previous_results}.
\begin{table}[h]
    \centering
    \resizebox{\columnwidth}{!}{
    \begin{tabular}{|l|c|c|}
        \hline
        Model & Accuracy  & Selection Rate \\
        \hline 
        \multirow{2}{*}{Baseline} &  73.28\% & 
        \multirow{2}{*}{49.63\%} \\
        & \textbf{97.94\%} / \textit{48.61\%} & \cr 
        \hline
        \multirow{2}{*}{Fine-tuned FF} & 87.54\% & \multirow{2}{*}{83.59\%} \\
        & \textbf{95.36\%} / \textit{79.71\%} & \cr
        \hline
        \multirow{2}{*}{Fine-tuned Mixed} & 95.17\% & \multirow{2}{*}{97.46\%} \\ 
        & \textbf{93.94\%} / \textit{96.39\%} & \\
        \hline
        \multirow{2}{*}{Fine-tuned FF + Mixed} & 96.22\%  &   \multirow{2}{*}{99.95\%} \\
        & \textbf{96.20\%} / \textit{96.24\%} & \cr
        \hline
    \end{tabular}
    }
    \caption{Impact of fine-tuning with ground-truth labels of FairFace and mixed datasets. The first line corresponds to the general accuracy. Accuracy values for male and female are in \textbf{bold} and in \textit{italic}, respectively. }
    \label{tab:previous_results}
\end{table}

\subsection{Semi-Supervised Learning}
Since ground-truth labels are often unavailable or expensive to obtain, we adopt pseudo-labeling to improve model performance using unlabeled data. A model initially trained on labeled data is used to generate high-confidence predictions (pseudo-labels), which are then treated as ground truth for further training. We apply this approach using the FairFace dataset and generate pseudo-labels via the FixMatch framework \cite{sohn2020fixmatchsimplifyingsemisupervisedlearning}, testing confidence thresholds $\epsilon \in \{0.1,\ 0.3,\ 0.6,\ 0.9\}$. 
\begin{table}[h]
    \centering
    \resizebox{\columnwidth}{!}{%
    \begin{tabular}{|l|c|c|}
        \hline
        Model & Accuracy  & Selection Rate \\
        \hline 
        \multirow{2}{*}{Baseline} &  73.28\% & \multirow{2}{*}{49.63\%} \\
        & \textbf{97.94\%} / \textit{48.61\%} & \cr
        \hline
        \multirow{2}{*}{$\epsilon$ = 0.1}  & 77.26\% & \multirow{2}{*}{58.66\% }\\
        & \textbf{97.39\%} / \textit{57.13\%} & \cr
        \hline
        \multirow{2}{*}{$\epsilon$ = 0.3} & 79.04\% & \multirow{2}{*}{61.93\%} \\ 
        & \textbf{97.62\%} / \textit{60.46\%} & \\
        \hline
        \multirow{2}{*}{$\epsilon$ = 0.6} & 78.80\%  &  \multirow{2}{*}{63.58\%} \\
        & \textbf{96.32\%} / \textit{61.29\%} & \cr
        \hline
        \multirow{2}{*}{$\epsilon$ = 0.9}  & 79.42\%  &   \multirow{2}{*}{69.94\%} \\
        & \textbf{93.46\%} / \textit{65.37\%} & \cr
        \hline
    \end{tabular}%
    }
    \caption{Performance of one iteration of pseudo-labeling with no pseudo-balancing (no PB) under varying confidence thresholds on the AAF dataset.}
    \label{tab:pseudo-labelling-initial-results}
\end{table}
\subsubsection{Pseudo-Balancing}

We define \textbf{pseudo-balancing} as the enforcement of equal class representation during pseudo-label selection in self-training. Unlike standard pseudo-labeling, which accepts all high-confidence predictions from an initial model, pseudo-balancing selectively subsamples these predictions to ensure demographic parity—e.g., an equal number of pseudo-male and pseudo-female samples. This prevents the model from reinforcing majority-class biases during retraining. In our implementation, we apply pseudo-balancing after each pseudo-labeling step by partitioning the confident predictions into gender groups and sampling an equal number from each before feeding them into the retraining phase. This strategy complements confidence-based selection by adding a fairness constraint to the sample distribution, ensuring balanced learning signals during iterative self-training.

\subsubsection{Iterative Self-Training}
\begin{figure}[h]
    \centering
    \includegraphics[width=0.95\linewidth]{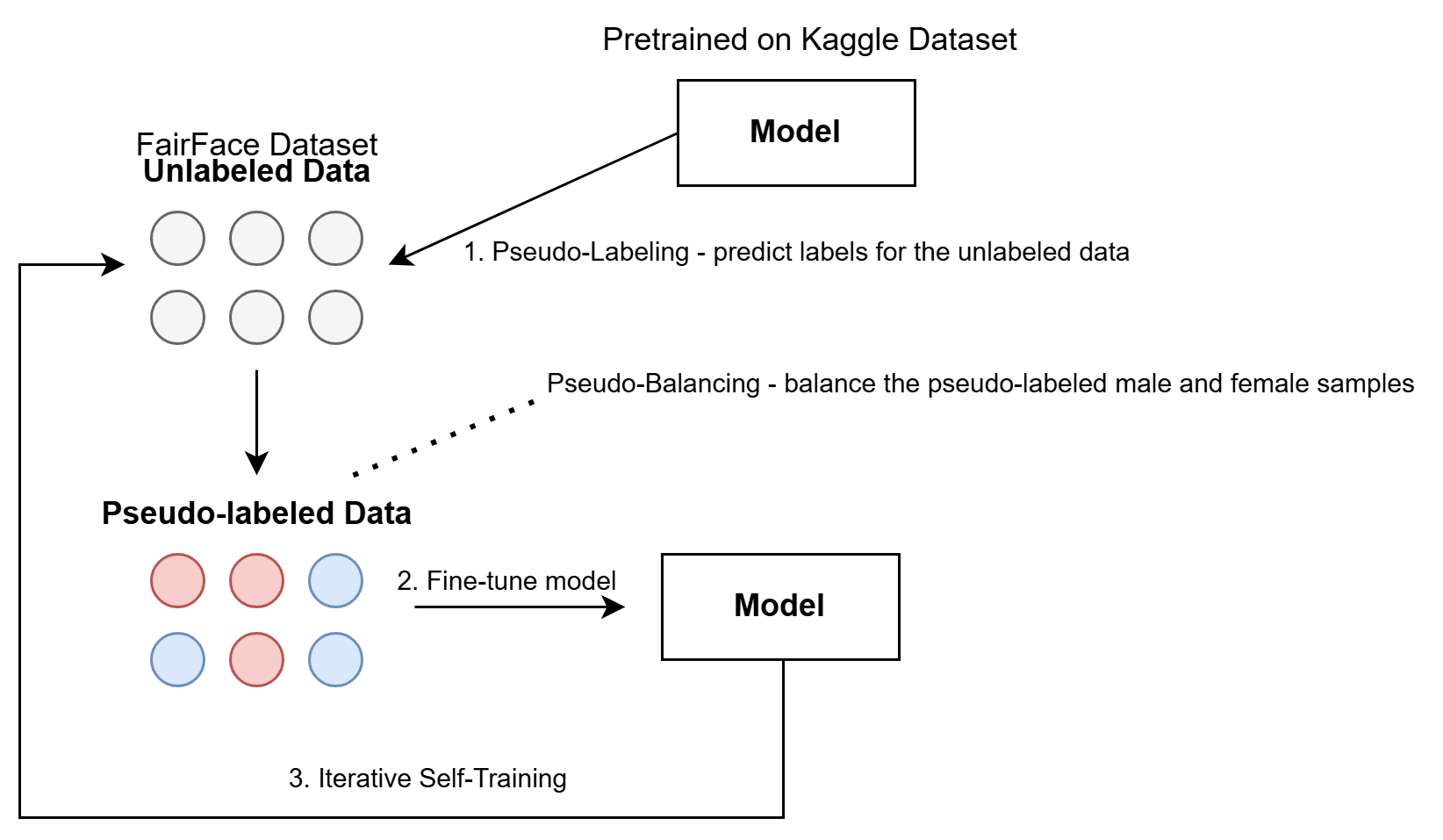}
    \caption{Iterative self-training using pseudo-labeled data.}
    \label{fig:iterative-self-training}
\end{figure}

One iteration of pseudo-labeling using a single-confidence threshold improves the performance of the baseline model (Table \ref{tab:pseudo-labelling-initial-results}). We extend single-step pseudo-labeling into an iterative self-training framework, where the model alternates between generating pseudo-labels for unlabeled data and retraining on these labels (Figure \ref{fig:iterative-self-training}). This approach has shown strong performance in both classification and segmentation tasks \cite{zou2018domainadaptationsemanticsegmentation, zoph2020rethinkingpretrainingselftraining, chen2020naivestudentleveragingsemisupervisedlearning, Feng2020SemiSupervisedSS, xie2020selftrainingnoisystudentimproves}.

We implement two variants of this strategy. FixMatch \cite{sohn2020fixmatchsimplifyingsemisupervisedlearning} combines weak and strong augmentations with confidence-based pseudo-label selection. Weak augmentations generate reliable pseudo-labels, while strong augmentations provide more robust training signals. After each step, we enforce demographic parity via \textbf{pseudo-balancing}, subsampling the pseudo-labeled set to maintain equal representation of male and female faces. FlexMatch \cite{zhang2022flexmatchboostingsemisupervisedlearning} builds on FixMatch by introducing a curriculum-based thresholding scheme that adapts confidence thresholds per class based on learning progress. This dynamic adjustment may improve learning for underrepresented classes, particularly female samples, and reduces bias amplification common in imbalanced datasets.

\section{Experiments}
\label{sec:experiments}
To assess how pseudo-labeling and pseudo-balancing strategies behave under skewed demographic conditions, we conducted a set of controlled experiments using the FairFace dataset. The experiments are structured around two key scenarios: 1) mitigating bias when using a balanced unlabeled dataset with a demographically biased initial model; and 2) analyzing the effects of demographic skew within the unlabeled training data. 

We split the FF dataset into 86,744 samples for training (40,758 female and 45,986 males) and 1,250 East-Asian samples (777 males, 773 females) for validation. For evaluation, a gender-balanced test set of 2,504 samples was curated from AAF.  To investigate bias patterns, we created specialized subsets of FF including a 43,372-sample gender-balanced variant (GB-FF), race-specific collections (12,287 East-Asian, 12,233 Black, and 6,141 East-Asian samples), and gender-skewed versions (41,803 samples with a 80\% female distribution and 44,941 samples with 80\% male distribution). 

\subsection{\textbf{Scenario 1: Bias Mitigation with Balanced Unlabeled Data}}

In the first scenario, we simulate a realistic setting where an off-the-shelf model exhibits inherent biases due to its training history. We use the pre-trained Kaggle CNN PyTorch model, which is known to perform poorly on minority subgroups, as our initial model. For fine-tuning, we employ unlabeled data from FairFace—specifically its race-balanced (FF) and gender-balanced (GB-FF) subsets. Although FairFace includes gender labels, these are used only for evaluation, not training. 

Evaluation is conducted on the AAF benchmark, which predominantly features East Asian individuals and allows for detailed demographic bias analysis. We compare the baseline Kaggle model against variants fine-tuned on FF and GB-FF using FixMatch and FlexMatch, with and without pseudo-balancing (PB).

\subsection{\textbf{Scenario 2: Learning with Biased Unlabeled Data}}

In this scenario, we investigate how pseudo-balancing performs when the unlabeled training data itself is biased. We design the following controlled bias conditions:
\begin{enumerate}
\item \textit{Gender Bias}: Datasets with 80\%-20\% or 50\%-50\% male-female distributions.
\item \textit{Severe Racial Bias}: Datasets containing only one race (e.g., East Asian or Black).
\item \textit{Gender and Racial Bias}: An East Asian female FairFace subset to simulate overlapping gender and racial bias, reflecting AAF test set characteristics.
\end{enumerate}
    
\subsection{Results}

FixMatch was evaluated using two confidence thresholds ($\epsilon$ = 0.6 and $\epsilon$ = 0.9) to analyze the impact of pseudo-labeling confidence on model performance. FlexMatch used a non-linear threshold adjustment based on class-specific learning status, with a base threshold of 0.95, in which the adjustment function is given by $\frac{x}{2-x}$. Training was conducted for 10 epochs per iteration using SGD optimization (learning rate = 0.001, momentum = 0.9) and a batch size of 16, with early stopping applied after 2 epochs of no improvement. These parameters are based on previous ablation studies of FlexMatch performance \cite{zhang2022flexmatchboostingsemisupervisedlearning}. The best models can be seen in Table \ref{tab:pseudo-labelling_fixmatch} and Table \ref{tab:pseudo-labelling_flexmatch}, highlighted in yellow.

\begin{table}[h]
    \centering
    \resizebox{\columnwidth}{!}{%
        \begin{tabular}{|l|c|c|c|c|}
            \hline
        Model & $\epsilon$ & PB & Accuracy & SR \\
            \hline 
        Baseline & - & - & 73.28\% & 49.63\% \\
            & & & \textbf{97.94} / \textit{48.61} & \\ 
             \hline
\colorbox{yellow}{FairFace} & 0.6 & yes &   80.05\% & 76.80\% \\
            & & & \textbf{91.12} / \textit{69.98} & \\ 
            \hline
FairFace & 0.6 & no & 76.92\% & 59.93\% \\
            & & & \textbf{96.20} / \textit{57.65} & \\ 
            \hline
            
\colorbox{yellow}{FairFace} & 0.9 & yes & 79.81\% & 93.80\% \\
            & & & \textbf{82.37} / \textit{77.26} & \\ 
            \hline     
            
FairFace & 0.9 & no & 75.73\% & 55.59\% \\
            & & & \textbf{97.35} / \textit{54.12} &  \\ 
            \hline



Gender-Balanced FF & 0.9 & yes & 78.65\% & 63.91\% \\
            & & & \textbf{95.96} / \textit{61.33} & \\ 
            \hline


            
Severe East-Asian Female FF & 0.9 & yes &  71.63\% & 48.82\% \\
            & & & \textbf{90.33} / \textit{44.10} &  \\ 
            \hline
            
\colorbox{yellow}{Severe East-Asian Female FF} & 0.9 & no &  81.30\% & 84.78\%  \\
            & & & \textbf{87.99} / \textit{74.60} & \\ 
            \hline



Severe Black FF & 0.9 & yes  &  74.41\% & 53.87\% \\
            & & & \textbf{96.72} / \textit{52.10} & \\ 
            \hline

Severe Black FF & 0.9 & no  &  71.35\% & 44.08\% \\
            & & & \textbf{99.05} / \textit{43.66} & \\ 
            \hline
\colorbox{yellow}{Severe Male FF} & 0.6 & yes &  78.11 \% & 80.96\%  \\
            & & & \textbf{86.33} / \textit{69.89} & \\ 
            \hline

 
            
Severe East-Asian FF & 0.6 & yes & 79.33\% & 76.36\%  \\
            & & & \textbf{90.65} / \textit{69.22} & \\ 
            \hline
            

Severe East-Asian FF & 0.9 & yes & 80.59\% & 68.66\%  \\
            & & & \textbf{95.56} / \textit{65.61} & \\ 
            \hline
            

            
Severe Female FF & 0.9 & no & 78.86\%  & 64.58\% \\
            & & & \textbf{95.84} / \textit{61.89} & \\ 
            \hline

        \end{tabular}%
}
    \caption{Performance of the fine-tuned model with the pseudo-labeled data from FairFace with different confidence scores of FixMatch with and without pseudo-balancing, evaluated on the AAF dataset. Initial model is the pre-trained Kaggle CNN PyTorch model. Accuracy values for male and female are in \textbf{bold} and in \textit{italic}, respectively.}
    \label{tab:pseudo-labelling_fixmatch}
\end{table}

\begin{table}[h]
    \centering
    \resizebox{\columnwidth}{!}{%
        \begin{tabular}{|l|c|c|c|c|}
            \hline
            Model & PB & Accuracy & SR \\
            \hline 
            Baseline & - & 73.28\% & 49.63\% \\
            & &  \textbf{97.94} / \textit{48.61} & \\ 
            \hline
\colorbox{yellow}{FairFace} & yes &  83.36\% & 86.03\% \\
            & & \textbf{89.62} / \textit{77.10} & \\ 
            \hline
\colorbox{yellow}{FairFace} & no &  79.18\% & 74.77\% \\
            & & \textbf{67.75} / \textit{90.61} & \\ 
            \hline
Gender-Balanced FF & yes &  67.49\% & 42.19\% \\
            & & \textbf{94.93} / \textit{40.05} &   \\ 
            \hline
            
Severe East-Asian Female FF & yes &  71.87\% & 72.27\% \\
            & & \textbf{83.44} / \textit{60.30} & \\ 
            \hline
Severe East-Asian Female FF & no &  56.04\% & 47.37\% \\
            & & \textbf{76.07} / \textit{36.01} & \\ 
            \hline
Severe Male FF & yes &  62.92\% & 71.86\% \\
            & & \textbf{52.61} / \textit{73.21}&  \\ 
            \hline
Severe Male FF & no &  54.16\% & 21.28\% \\
            & & \textbf{19.02} / \textit{89.30} & \\ 
            \hline
        \end{tabular}%
    }
    \caption{Performance of the fine-tuned model with the pseudo-labeled data from FairFace with FlexMatch, evaluated on the AAF dataset. Initial model is the pre-trained Kaggle CNN PyTorch model. Accuracy values for male and female are in \textbf{bold} and in \textit{italic}, respectively.}
    \label{tab:pseudo-labelling_flexmatch}
\end{table}

\begin{figure}[t]
    \centering
    \includegraphics[width=0.95\linewidth]{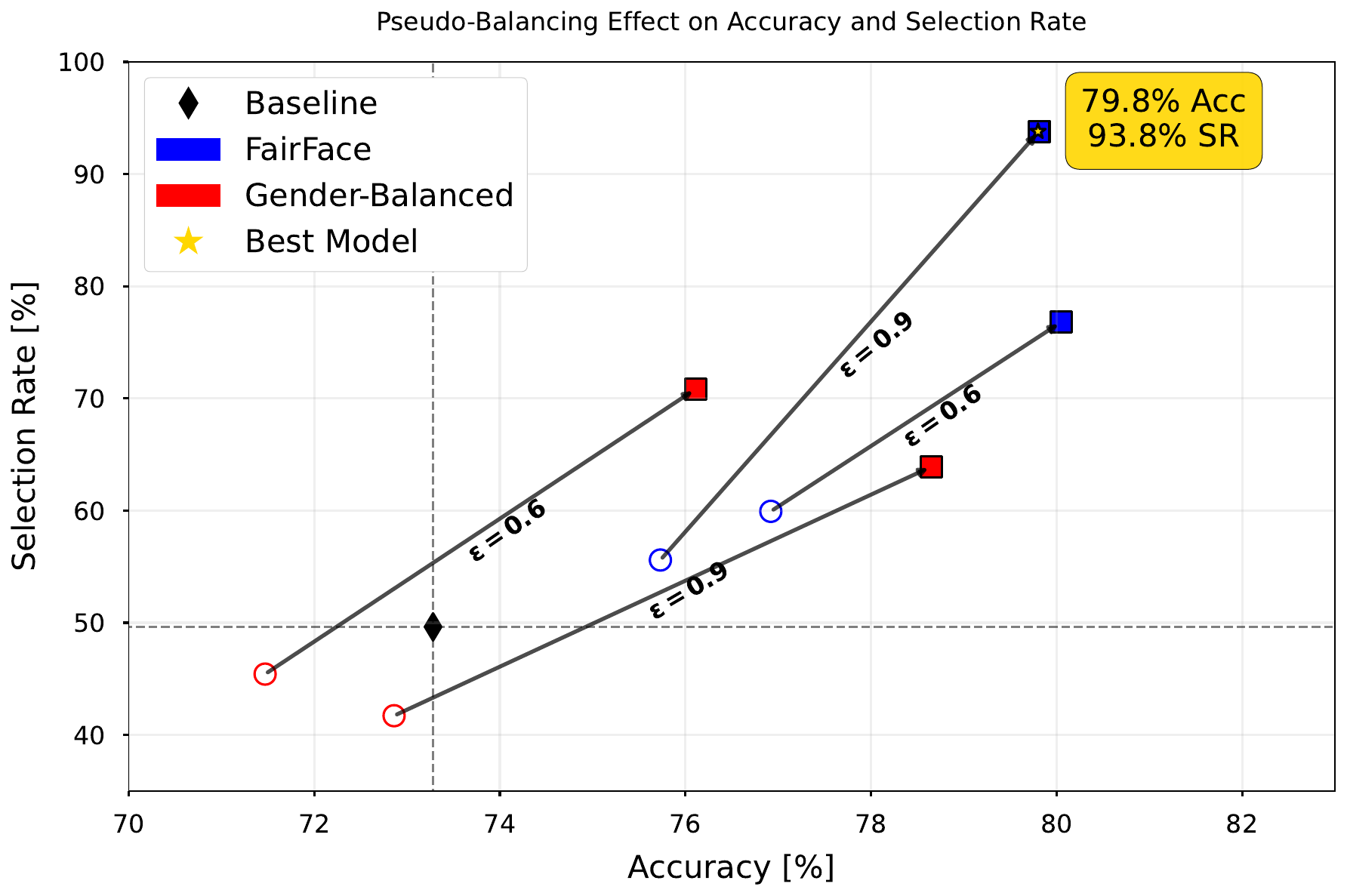}
    \caption{Scenario 1. Accuracy vs. selection rate for FF variants using FixMatch. Squares denote pseudo-balanced (PB) results and circles non-PB counterparts. Arrows indicate performance shifts. The best model achieves 82.4\% accuracy at 93.8\% selection rate (highlighted).}
    \label{fig:fixmatch_baseline}
\end{figure}

\begin{figure}[t]
    \centering
    \includegraphics[width=0.95\linewidth]{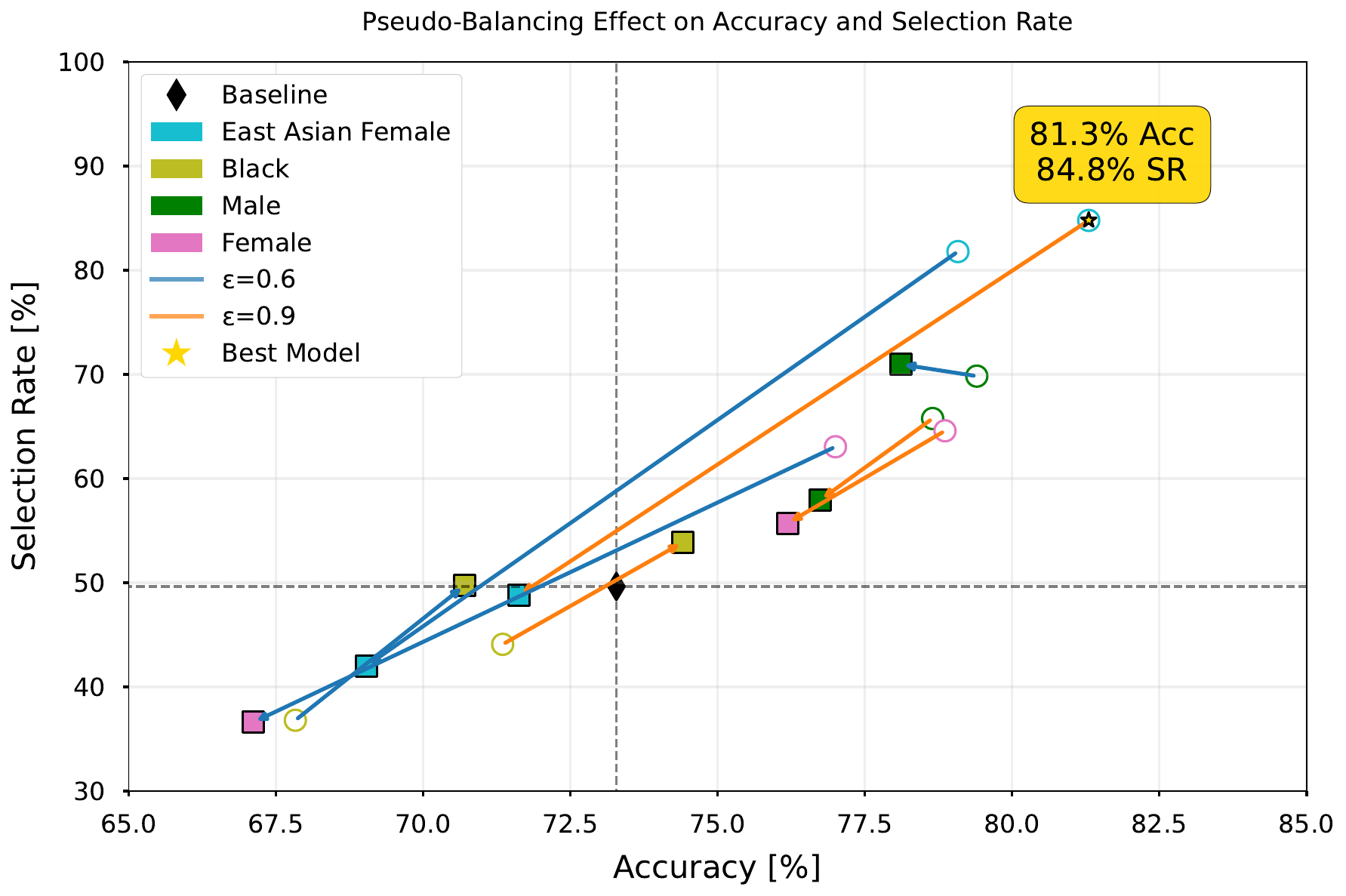}
    \caption{Scenario 2. Bias-specific performance analysis showing East Asian Female, Black and Male subsets using FixMatch. Squares denote pseudo-balanced results and circles show non-PB results ($\epsilon=0.6$: blue arrows; $\epsilon=0.9$: orange arrows). The model trained on the East Asian Female subset achieves 81.3\% accuracy at 84.8\% selection rate (highlighted).}
    \label{fig:fixmatch_biased}
\end{figure}

\begin{figure}[t]
    \centering
    \includegraphics[width=0.95\linewidth]{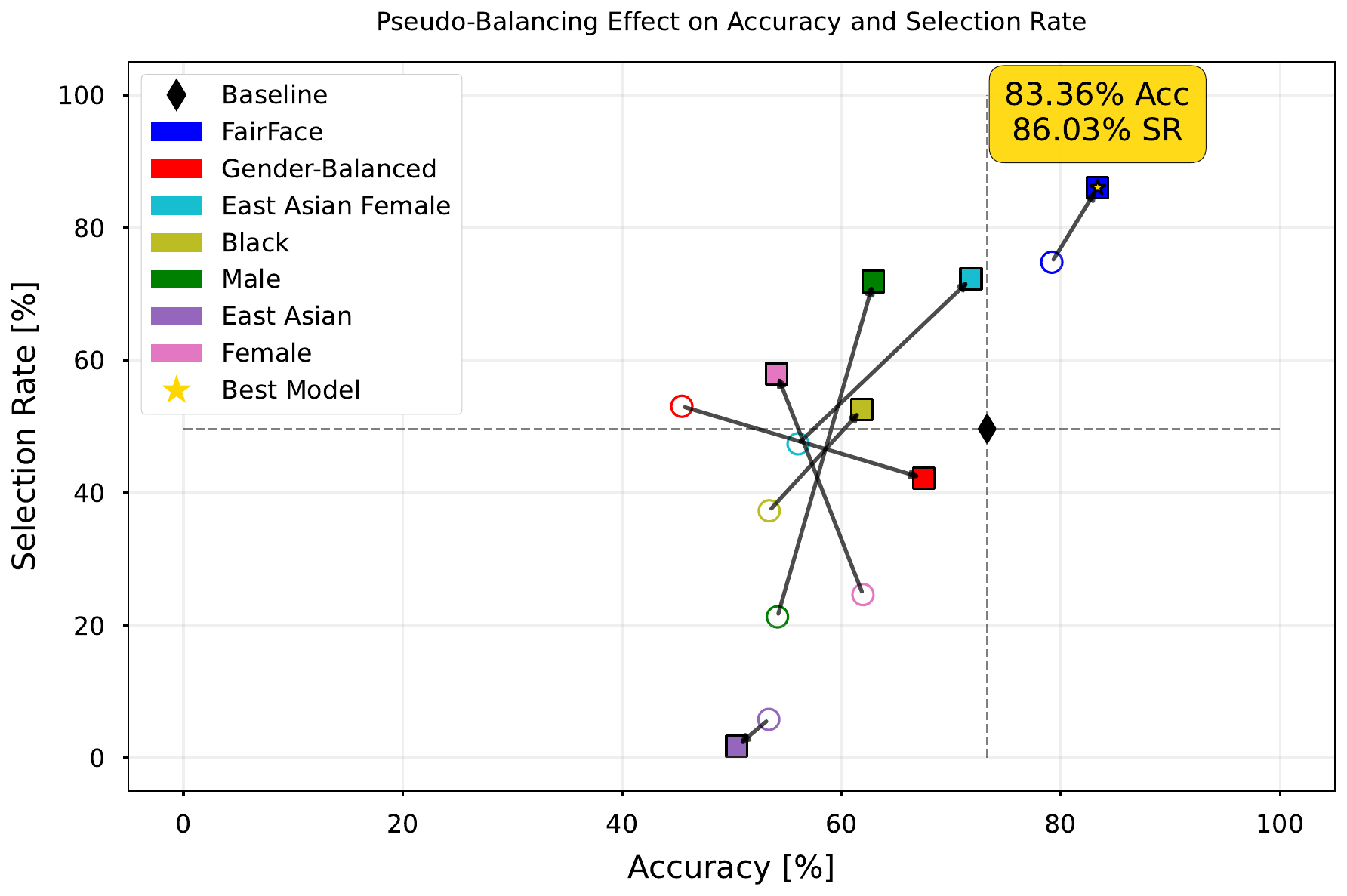}
    \caption{Scenarios 1 and 2. FlexMatch adaptation results across all bias conditions. Squares denote PB results and circles non-PB results. The best model achieves 83.36\% accuracy at 86.03\% selection rate using FairFace dataset with PB.}
    \label{fig:flexmatch_results}
\end{figure}

Pseudo-balancing (PB) improves performance in most cases, particularly for datasets that are initially balanced or diverse like FF and GB-FF (Fig. \ref{fig:fixmatch_baseline}). For instance, FairFace with PB achieves 83.36\% accuracy with FlexMatch, compared to 79.18\% without PB; FixMatch ($\epsilon$ = 0.9, PB) achieves a accuracy of 79.81\% and 75.73\% ($\epsilon$ = 0.9, No PB). On severely biased datasets like Severe East-Asian Female and Severe Male datasets, pseudo-balancing may degrade model performance (Fig. \ref{fig:fixmatch_biased}). For example, on the Severe East-Asian Female dataset, FixMatch ($\epsilon$ = 0.9, PB) achieves 81.30\% accuracy, compared to 71.63\% with PB. This suggests that pseudo-balancing is most effective when applied to training datasets that are initially diverse or balanced, as it may help maintain class equilibrium during training and prevent the model from overfitting to dominant classes. However, in cases where the training dataset exhibits severe biases, pseudo-balancing can inadvertently amplify existing imbalances, leading to accelerated bias. This occurs because pseudo-balancing may reinforce the skewed distribution of pseudo-labels, further disadvantaging underrepresented classes. This observation is supported by \cite{lee2013pseudo}, which emphasizes the importance of maintenance of dataset composition in pseudo-labeling strategies.

FlexMatch achieves its best performance of 83.36\% accuracy (with a SR of 86.03\%) when trained with FF and pseudo-balancing. However, its effectiveness diminishes significantly when dealing with severely biased datasets, where it underperforms compared to both FixMatch and the baseline model. This suggests that adaptive thresholding methods, while effective in balanced or moderately biased scenarios, are insufficient to address extreme gender or racial biases without additional balancing mechanisms such as pseudo-balancing. 

Furthermore, the presence of severe bias in the training data significantly impacts model performance on the AAF dataset. When fine-tuning the model with FixMatch using severely biased datasets such as the East-Asian FairFace or East-Asian Female FairFace, the model achieves accuracies of 79.33\% (with a SR of 76.36\%) and 81.30\% (with a SR of 84.78\%), respectively. These results demonstrate that models fine-tuned on datasets with specific racial or gender biases can perform on par with those fine-tuned on the entire FF or FF-GB dataset, provided the training data aligns with the target domain's demographic characteristics. Moreover, the Severe Male FF dataset performs well under FixMatch ($\epsilon = 0.6$, PB). In this setting, pseudo-balancing effectively compensates for extreme gender bias by redistributing pseudo-labels, while the reduced threshold helps preserve gender diversity and enhances recall for the underrepresented female class.

\subsection{Self-Training Performance Analysis}

\begin{figure*}[h]
    \centering
    \begin{subfigure}[b]{0.45\textwidth}
        \centering
        \includegraphics[width=\textwidth]{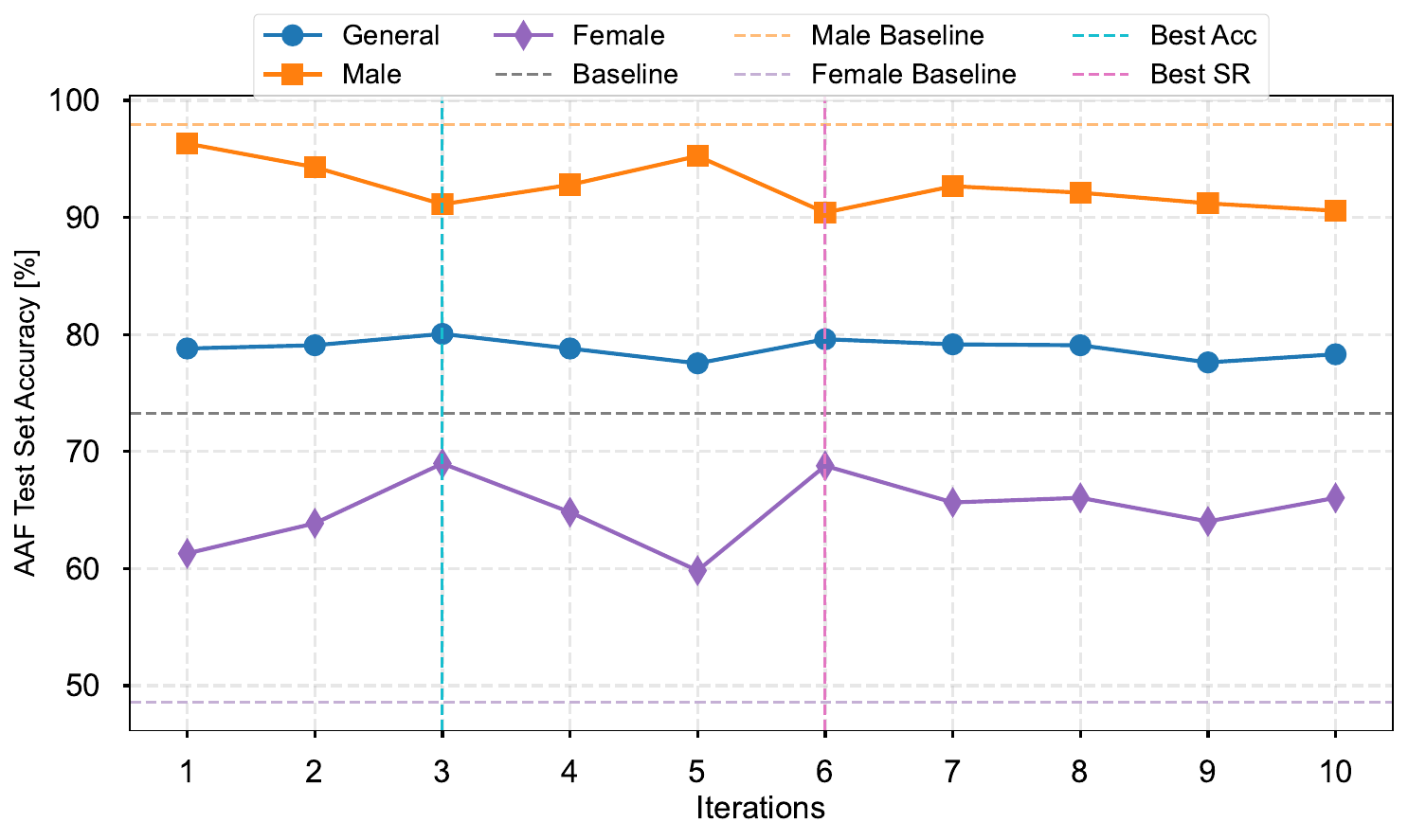}
        \caption{FixMatch with $\epsilon$ = 0.6 with pseudo-balancing}
    \end{subfigure}
    \hfill
    \begin{subfigure}[b]{0.45\textwidth}
        \centering
        \includegraphics[width=\textwidth]{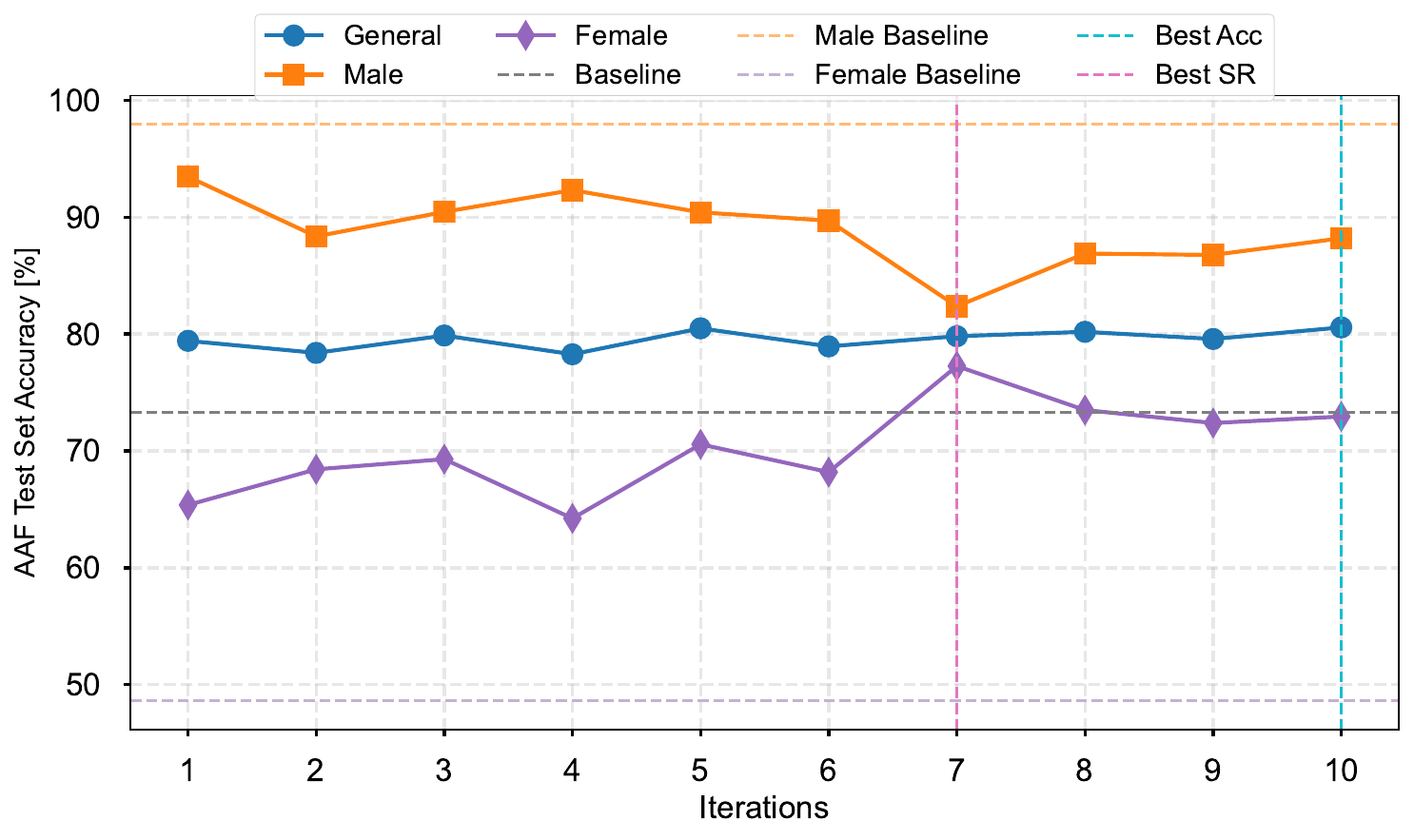}
        \caption{FixMatch with $\epsilon$ = 0.9 with pseudo-balancing}
    \end{subfigure}
    
    \vskip 0.2cm 

    \begin{subfigure}[b]{0.45\textwidth}
        \centering
        \includegraphics[width=\textwidth]{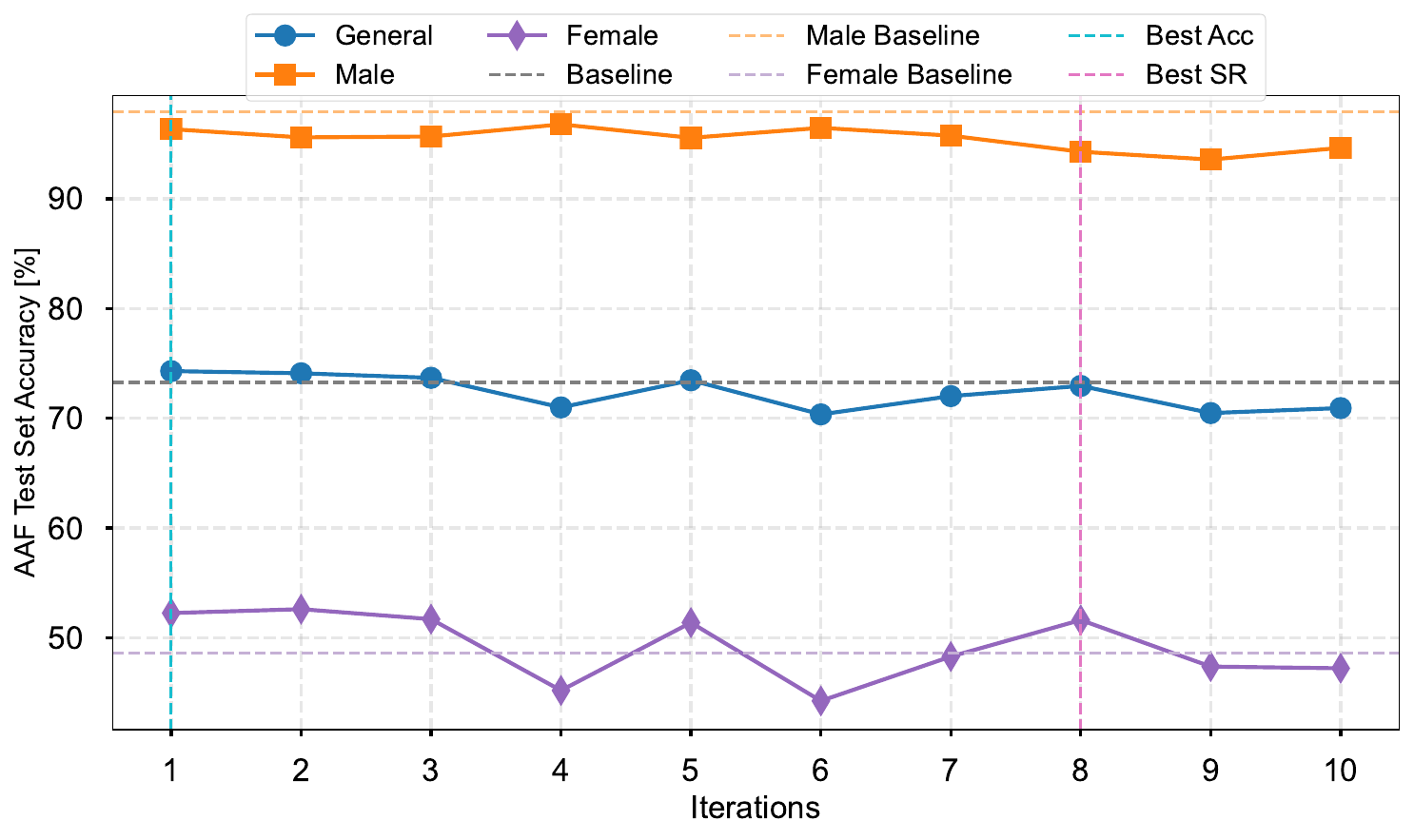}
        \caption{FixMatch with $\epsilon$ = 0.6 without pseudo-balancing}
    \end{subfigure}
    \hfill
    \begin{subfigure}[b]{0.45\textwidth}
        \centering
        \includegraphics[width=\textwidth]{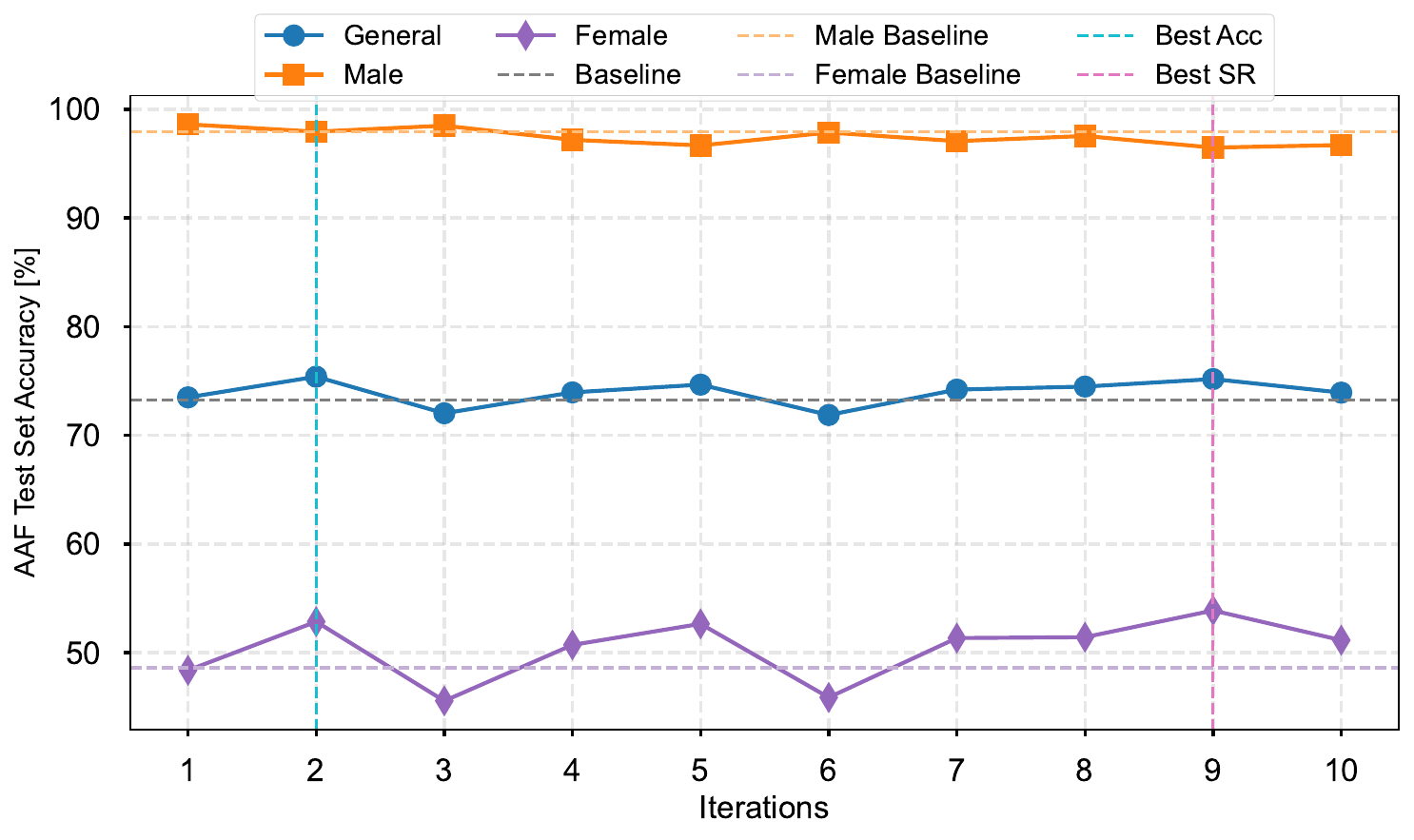}
        \caption{FixMatch with $\epsilon$ = 0.9 without pseudo-balancing}
    \end{subfigure}

    \caption{Scenario 1. Self-training per-iteration evolution of fine-tuned models using FixMatch, trained with unlabeled FF and tested on AAF dataset.}
    \label{fig:aaf-iterative-training}
\end{figure*}

\begin{figure}[h]
    \centering
    \begin{subfigure}[b]{0.23\textwidth}
        \centering
        \includegraphics[width=\textwidth]{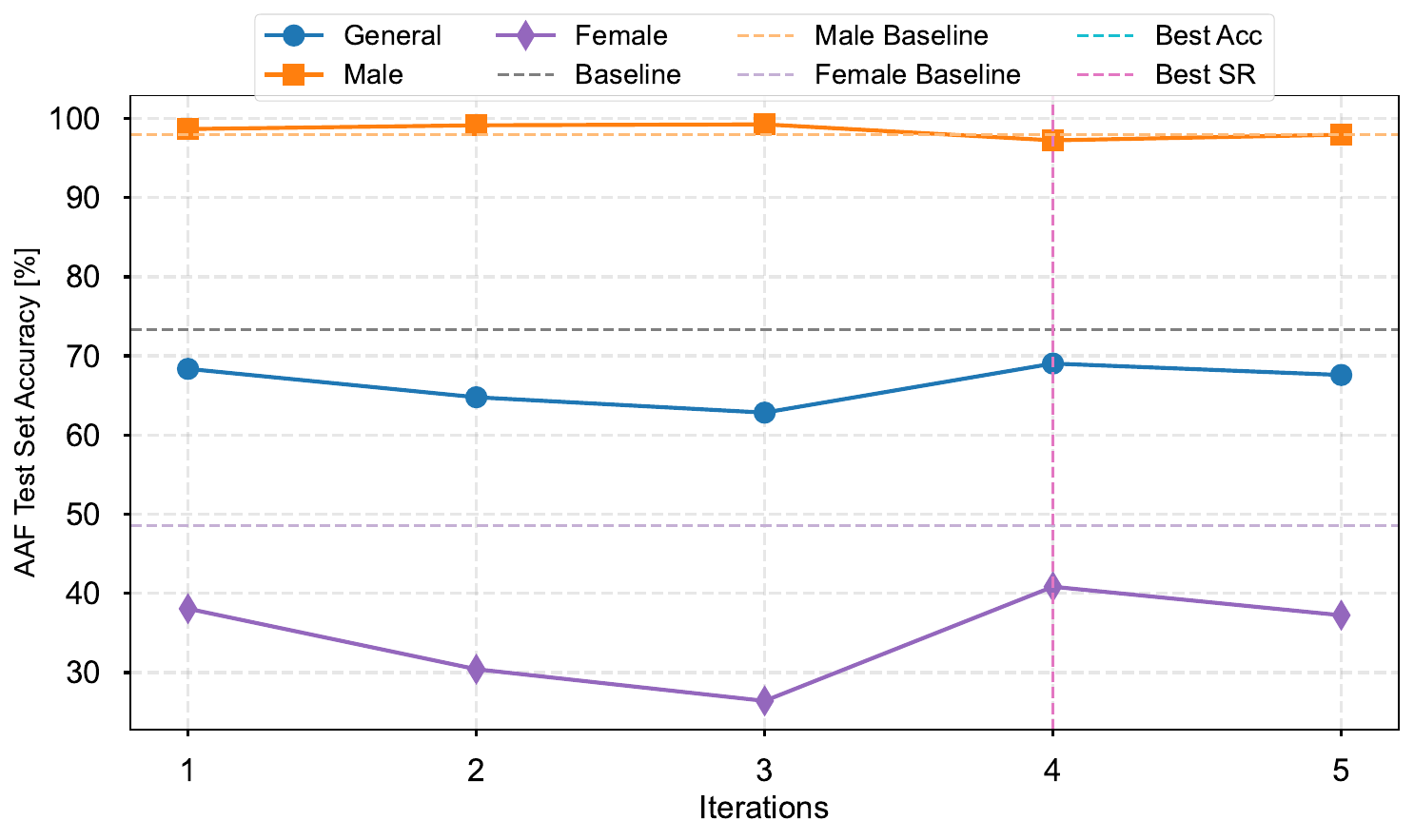}
        \caption{$\epsilon$ = 0.6 with PB}
    \end{subfigure}
    \hfill
    \begin{subfigure}[b]{0.23\textwidth}
        \centering
        \includegraphics[width=\textwidth]{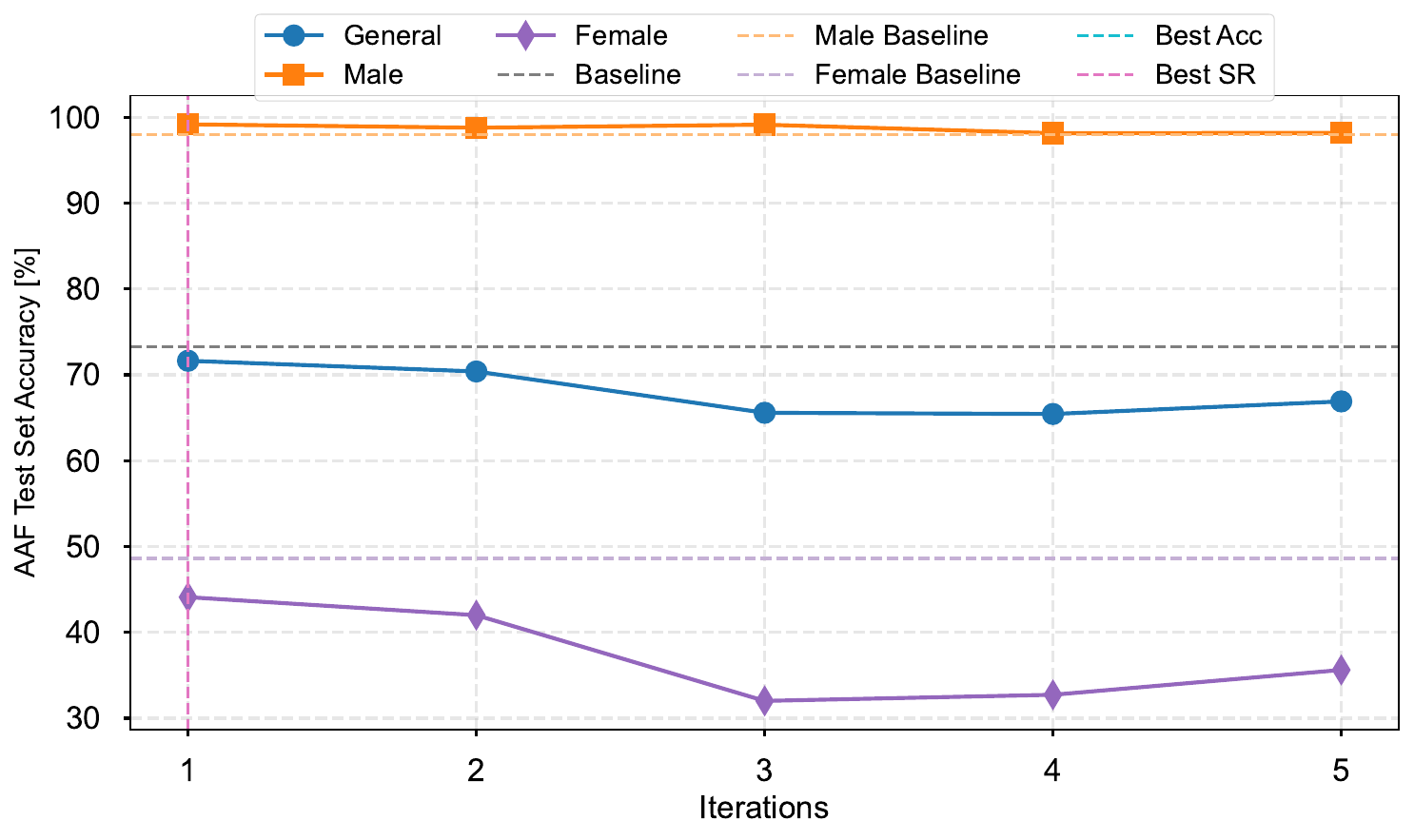}
        \caption{$\epsilon$ = 0.9 with PB}
    \end{subfigure}
    
    \vskip 0.3cm 

    \begin{subfigure}[b]{0.23\textwidth}
        \centering
        \includegraphics[width=\textwidth]{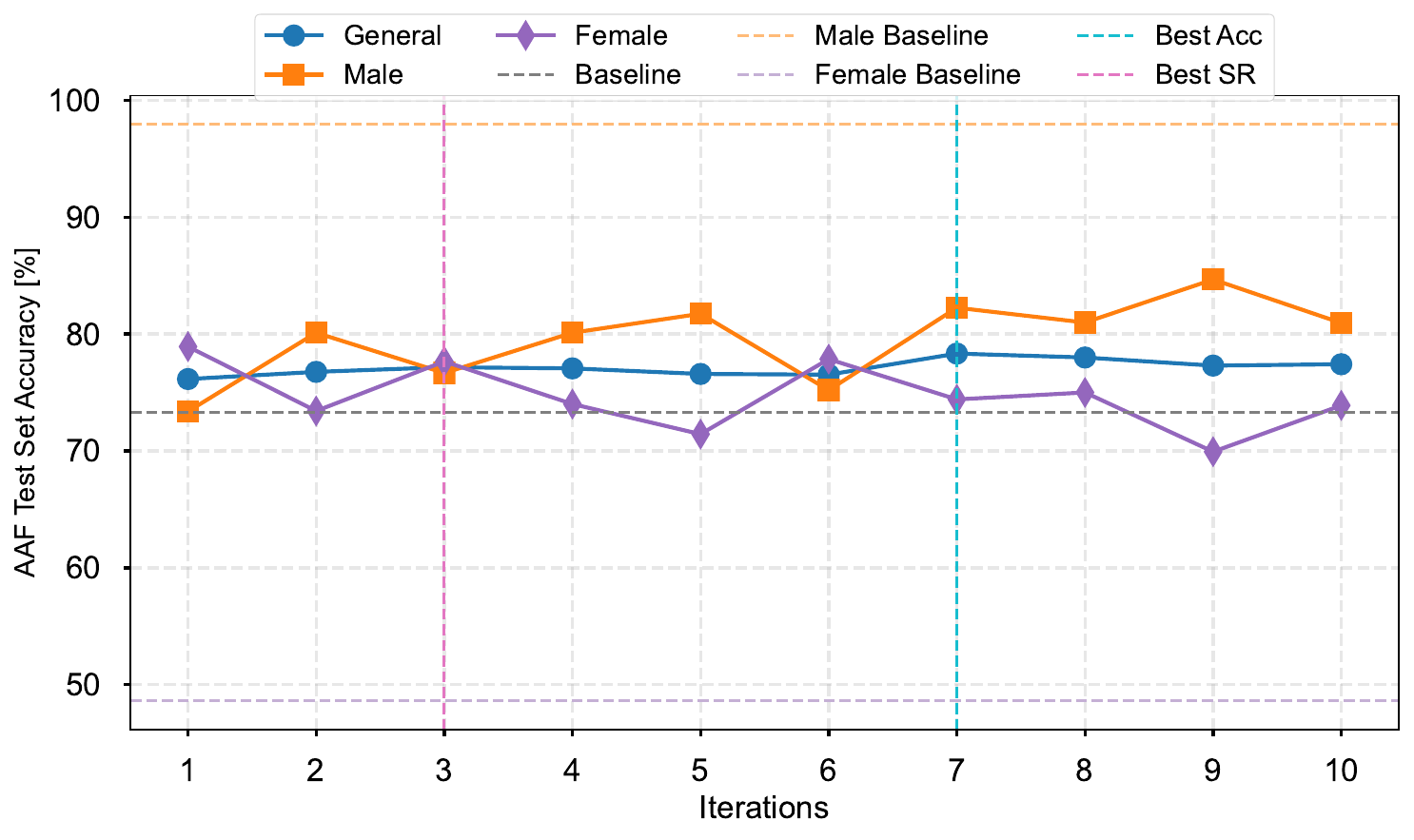}
        \caption{$\epsilon$ = 0.6 without PB}
    \end{subfigure}
    \hfill
    \begin{subfigure}[b]{0.23\textwidth}
        \centering
        \includegraphics[width=\textwidth]{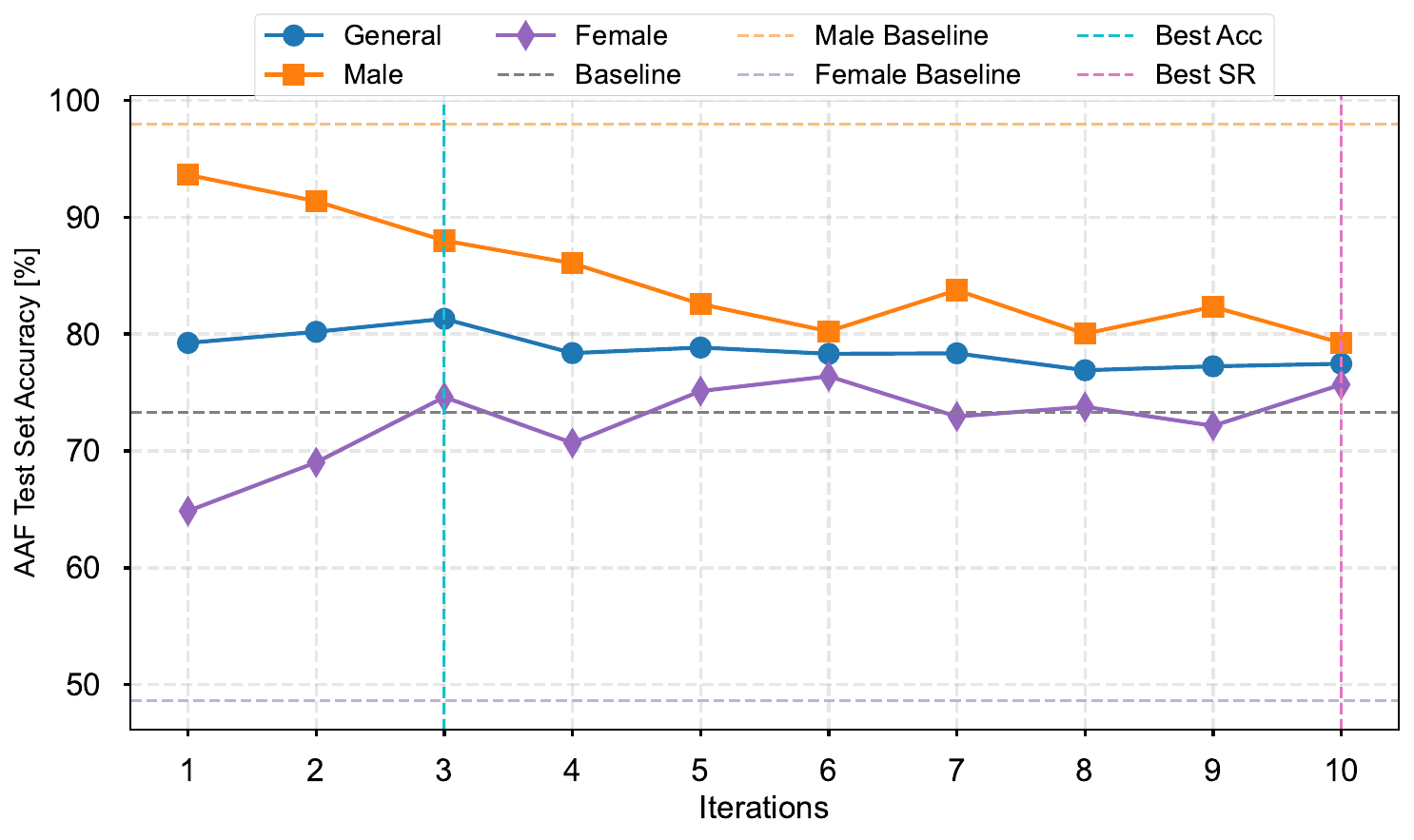}
        \caption{$\epsilon$ = 0.9 without PB}
    \end{subfigure}

    \caption{Self-training per-iteration evolution of fine-tuned models using FixMatch, trained with unlabeled East Asian Female biased FF and tested on AAF dataset.}
    \label{fig:aaf-iterative-training-fixmatch-east-asian}
\end{figure}

The iterative self-training results of FixMatch and FlexMatch highlight critical differences in bias mitigation efficacy and training stability for the current classification task (Fig. \ref{fig:aaf-iterative-training} and \ref{fig:aaf-iterative-training-fixmatch-east-asian}). While both methods achieve comparable overall accuracy (around 80\%), FixMatch ($\epsilon$ = 0.9, PB) reduces the gender accuracy gap to 5.01\% by iteration 7, surpassing FlexMatch by 7.51\%. This aligns with \cite{sohn2020fixmatchsimplifyingsemisupervisedlearning}, demonstrating that higher confidence thresholds enhance pseudo-label quality by filtering uncertain predictions, thereby mitigating bias propagation during training. The synergy of pseudo-balancing and elevated confidence thresholds prevents overfitting to dominant classes, allowing fairer gender performance. 

\begin{figure}[h]
    \centering
    \begin{subfigure}[b]{0.23\textwidth}
        \centering
        \includegraphics[width=\textwidth]{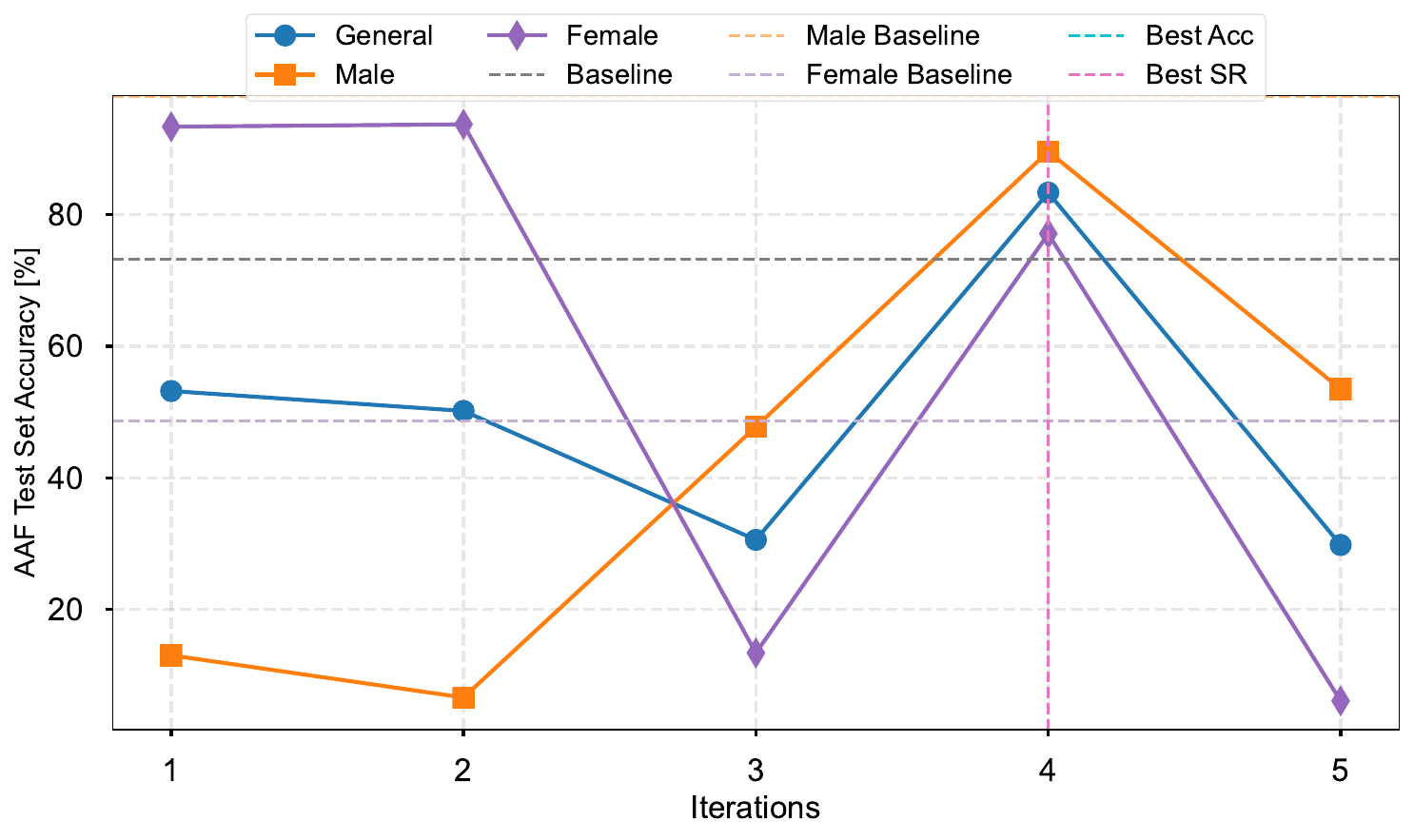}
        \caption{FlexMatch FF with PB}
    \end{subfigure}
    \hfill
    \begin{subfigure}[b]{0.23\textwidth}
        \centering
        \includegraphics[width=\textwidth]{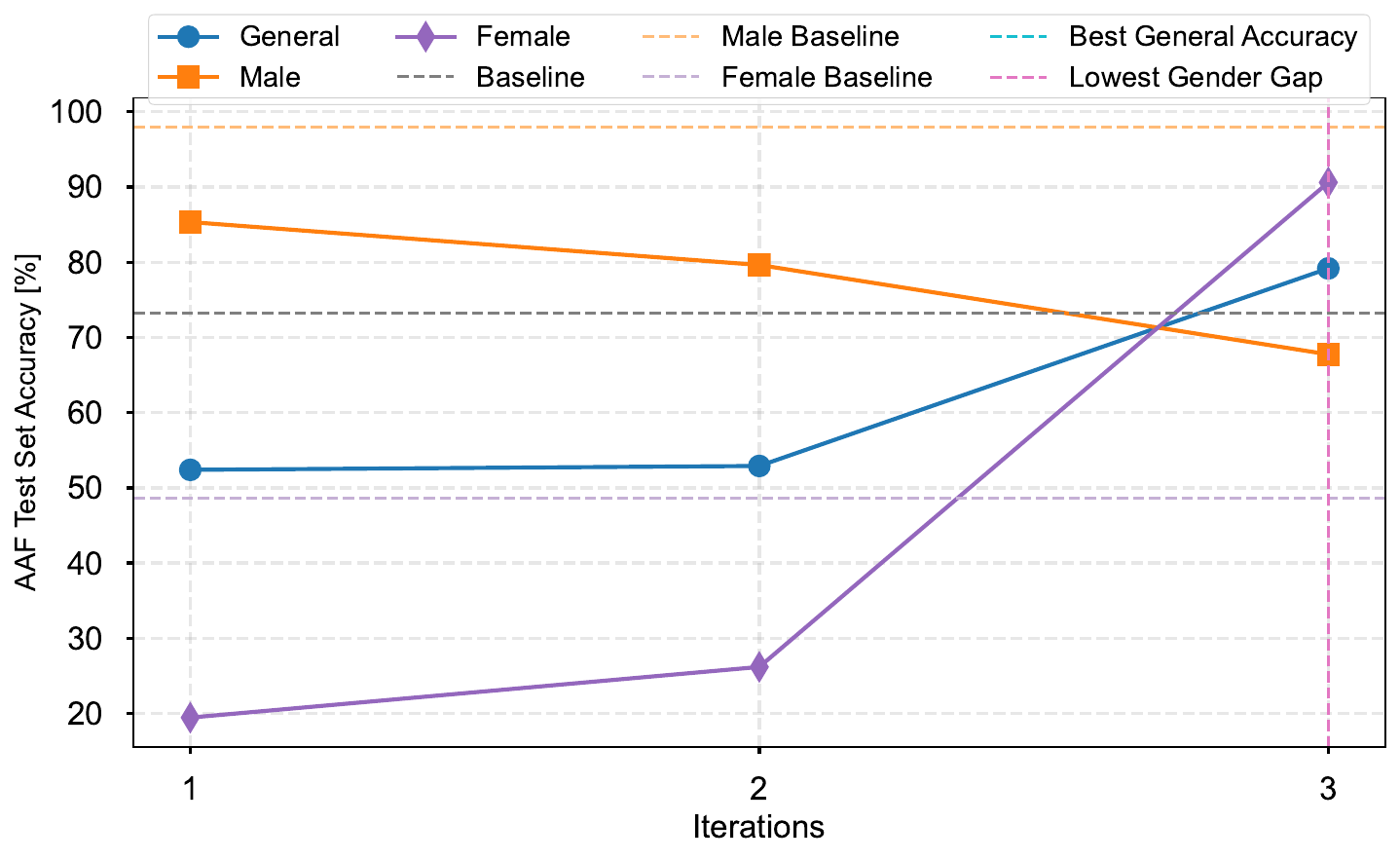}
        \caption{FlexMatch FF without PB}
    \end{subfigure}
    
    \vskip 0.3cm 

    \begin{subfigure}[b]{0.23\textwidth}
        \centering
        \includegraphics[width=\textwidth]{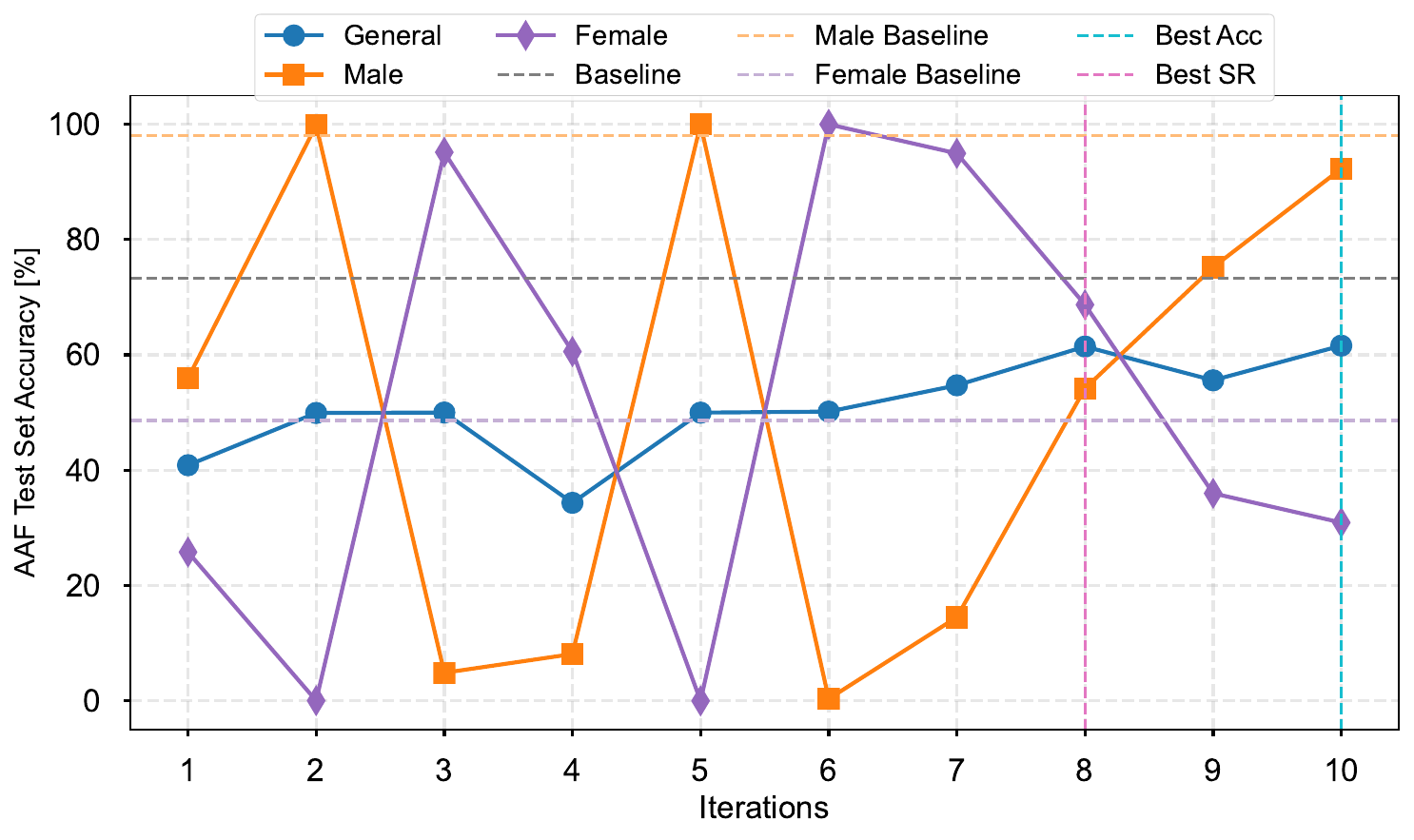}
        \caption{FlexMatch East Asian Female with PB}
    \end{subfigure}
    \hfill
    \begin{subfigure}[b]{0.23\textwidth}
        \centering
        \includegraphics[width=\textwidth]{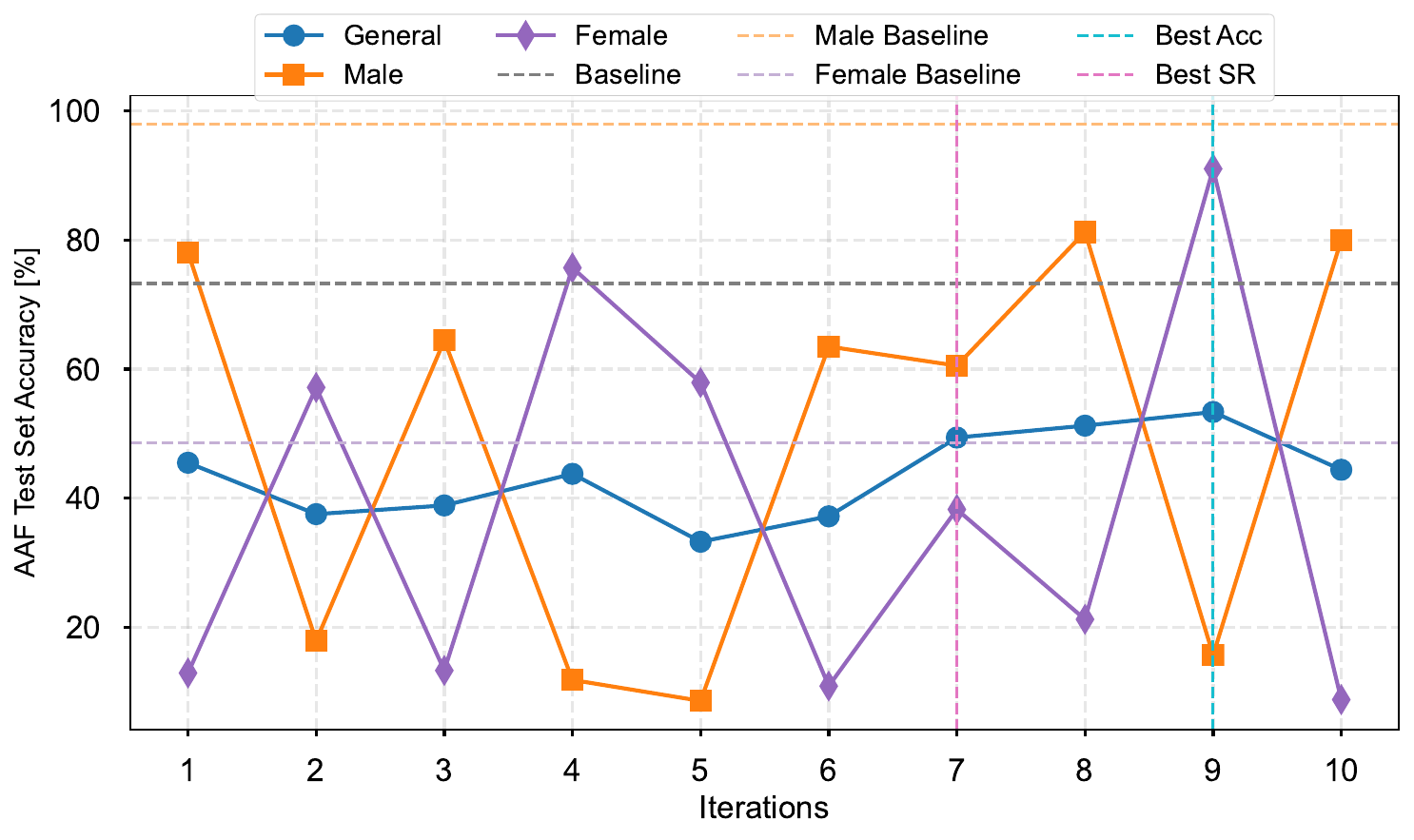}
        \caption{FlexMatch East Asian Female without PB}
    \end{subfigure}

    \caption{Self-training per-iteration evolution of fine-tuned models using FlexMatch, trained with unlabeled FF and East Asian Female biased FF and tested on AAF dataset.}
    \label{fig:aaf-iterative-training-flex}
\end{figure}

FlexMatch achieves the best overall accuracy (83.36\%) with FF dataset but exhibits instability when fine-tuning on severely biased distributions. The adaptive thresholding mechanism \cite{zhang2022flexmatchboostingsemisupervisedlearning} fails to stabilize training under extreme racial or gender imbalances (e.g., Severe East-Asian Female), resulting in erratic accuracy fluctuations (See Fig. \ref{fig:aaf-iterative-training-flex} and Table \ref{tab:pseudo-labelling_flexmatch}). Pseudo-balancing improves FlexMatch's performance by (1) compensating for FlexMatch’s tendency to ignore underrepresented classes early in training, and (2) stabilizing threshold adjustments when class-wise learning status is unreliable (See Fig. \ref{fig:flexmatch_results}). This explains why PB improves FlexMatch’s accuracy on GB-FF (67.49\% → 83.36\%) but fails on Severe East-Asian Female (71.87\% vs. no PB 56.04\%)—the baseline model’s poor initial performance on East Asian faces skews pseudo-label quality irrecoverably.

\section{Discussion}
\label{sec:discussion}

Our experiments reveal that pseudo-balancing effectively mitigates bias in semi-supervised learning when applied to balanced or moderately biased datasets, improving both accuracy and fairness. However, its efficacy diminishes when severe dataset biases align with the baseline model's intrinsic biases—particularly for East Asian subgroups, where compounding biases degrade performance. The Black subset benefits from partial alignment with the Western-centric pre-training data, highlighting how pseudo-balancing's success depends on the interplay between dataset composition, model bias, and target domain distribution. FlexMatch excels on balanced data but struggles with severe biases, underscoring the need for adaptive thresholding that accounts for these demographic imbalances. 

For race-balanced datasets like FairFace, pseudo-balancing mitigates the ''Matthew effect''—where the model reinforces existing biases by generating more confident pseudo-labels for already well-performing groups—by enforcing proportional sampling during self-training \cite{chen2022debiasedselftrainingsemisupervisedlearning}. This is particularly effective at higher confidence thresholds ($\epsilon$=0.9), where the 93.80\% selection rate indicates near-complete utilization of reliable pseudo-labels while maintaining a accuracy of 79.81\%. However, the East Asian performance reveals a critical limitation: when the baseline model's architectural bias (from Kaggle dataset pre-training) compounds with dataset bias, pseudo-balancing inadvertently reinforces incorrect feature representations. This manifests most severely in the East Asian Female subset, where pseudo-balancing amplifies the baseline model's accuracy deficit between male and female classes. The relative success of the Black FF subset (SR: 53.87\% with PB, 44.08\% without PB) indicates that the Kaggle model's Western-centric training provides better initial representations for African facial features than Asian ones. This difference likely arises because Western-centric datasets used for pre-training contain more diverse representations of African-descended faces than East Asian ones \cite{gender_shades_2018, mehrabi2019survey}, leading to better feature extraction for Black faces.

\section{Limitations and Future Work}
\label{sec:limitations}

The current study focuses primarily on gender and racial biases in facial analysis, though we recognize that real-world applications face additional challenges in age, expression, and cultural context variations \cite{wang2020mitigating}. The effectiveness of pseudo-balancing depends on the initial model's capability to generate reasonable pseudo-labels -- a limitation when dealing with completely novel demographic groups not represented in the pre-training data. Future work should investigate hybrid approaches combining our pseudo-balancing technique with demographic-aware augmentation strategies to handle more complex intersectional biases \cite{dooley2023rethinkingbiasmitigationfairer}. The promising results on East Asian populations implies that our method could be extended to other underrepresented groups, though this requires validation through larger-scale multicultural studies. 

\section{Conclusion}
\label{sec:conclusion}

We presented a simple yet effective method for mitigating gender bias in face gender classification through semi-supervised learning. Our approach, \textit{pseudo-balancing}, enforces demographic parity during pseudo-label selection without relying on any ground-truth demographic annotations. By leveraging demographically balanced pools of unlabeled data—such as the FairFace dataset—we guide biased models toward fairer outcomes. 

Through controlled experiments under both realistic and synthetic bias conditions, we demonstrate that pseudo-balancing improves classification accuracy while significantly reducing gender disparities, particularly on the All-Age-Faces benchmark, which predominantly features East Asian individuals. Although its effectiveness diminishes under severe data imbalance, our results reveal a key insight: access to a balanced or moderately skewed unlabeled dataset, combined with simple balancing constraints, can serve as a practical and scalable strategy for debiasing existing computer vision models.

These findings highlight pseudo-balancing as a simple and practical fairness intervention, especially in settings where demographic labels are unavailable. It offers a general framework for improving fairness in semi-supervised learning across broader computer vision tasks.

\section*{Acknowledgment}

This work was developed under the NII International Internship Program at National Institute of Informatics (NII) in Tokyo, Japan.

{
    \small
    \bibliographystyle{ieeenat_fullname}

}

\end{document}